\documentclass[11pt]{article}

\usepackage[final]{acl}

\usepackage{times}
\usepackage{latexsym}
\usepackage{amsmath}
\usepackage{amssymb}

\usepackage[T1]{fontenc}

\usepackage[utf8]{inputenc}

\usepackage{microtype}

\usepackage{inconsolata}

\usepackage{graphicx}

\usepackage{tabularx}
\usepackage{subcaption}
\usepackage{booktabs}

\title{Do LLMs Experience an Internal Polylogue?\\Investigating Reasoning through the Lens of Personas}

\author{
 \textbf{Nils A. Herrmann} \And
 \textbf{Leander Girrbach} \And
 \textbf{Kirill Bykov} \And
 \textbf{Zeynep Akata} \AND
 \normalfont Technical University of Munich, Helmholtz Munich and Munich Center for Machine Learning
\\
 \small{
   \textbf{Correspondence:} \href{mailto:nils.herrmann@tum.de}{nils.herrmann@tum.de}
 }
}

\begin{document}
\maketitle

\begin{abstract}
Recent work shows that large language models (LLMs) encode behavioral traits (``personas'') as linear directions in activation space, often called ``persona vectors''. Prior work has used such directions as static handles for behavioral steering. We instead treat them as dynamic signals: probes we can monitor and intervene on as reasoning unfolds. We use the term \textit{polylogue} to denote the time series of alignments between persona vectors and hidden activations over the course of generation. Experiments across four open-weight models show that polylogue features contain predictive signal for correctness comparable to low-dimensional activation summaries, while remaining interpretable through their associated persona directions. They also suggest concrete steering targets, namely, which latent directions to modulate at different stages of a response. We instantiate this as a simple paragraph-conditioned intervention that improves accuracy on three of the four models but degrades the fourth, suggesting that stage-aware latent steering is possible but not yet robust. Together, this positions the polylogue as an interpretable tool for reasoning-time monitoring and intervention.\footnote{
Code: \url{https://anonymous.4open.science/r/polylogue}}
\end{abstract}

\section{Introduction \& Related Work}

Multi-step reasoning improves LLM performance on complex tasks~\citep{weiChainofThoughtPromptingElicits2022}. Recent models explicitly incentivize longer-horizon behaviors such as reflection, planning, and systematic verification~\cite{guoDeepSeekR1IncentivizesReasoning2025, openaiOpenAIO1System2026}. A defining feature of the resulting reasoning traces is their \emph{temporal structure}: reasoning unfolds over time and engages distinct functional modes — problem comprehension, planning, exploration, verification of intermediate steps, and commitment to a final answer. Characterizing this structure and connecting it to model behavior is a central question for reasoning research.

We propose to study reasoning through a latent-space lens we call \emph{polylogue}: the time-varying engagement of multiple personas during a single chain-of-thought (CoT). Two recent threads of work motivate this view. First, behavioral traits, including persona-like styles, are encoded as linear directions in activation space and can be extracted via contrastive activation differences~\cite{subramaniExtractingLatentSteering2022, chenPersonaVectorsMonitoring2025}. Second, conceptual accounts argue that LLMs simulate a range of characters during pre-training, with post-training shaping one of these into the Assistant persona users interact with~\cite{PersonaSelectionModel}. Together, these results motivate testing whether a single CoT exhibits structured, time-varying alignment with persona-related activation directions. Building on this, we treat persona vectors not as static handles for behavioral steering, but as \emph{dynamic probes} that we can monitor and intervene on as reasoning unfolds. We use \emph{polylogue} purely as shorthand for this multivariate signal; it does not imply that the model instantiates discrete agents or voices.

Our work bridges two strands of reasoning research. The first analyzes CoT as a structured surface object, e.g.\ through self-consistency, step-wise revision, and verification-based selection~\citep{uesatoSolvingMathWord2022, lightmanLetsVerifyStep2023, jiangWhatMakesGood2025, xiongMappingMindsLLMs2025}, or by decomposing traces into functional reasoning episodes such as \emph{Analyze}, \emph{Explore}, and \emph{Verify}~\citep{liSchoenfeldsAnatomyMathematical2025,liUnderstandingThinkingProcess2025}, building on cognitive frameworks for human problem-solving~\citep{schoenfeldMathematicalProblemSolving1985}. These methods reveal informative patterns but are confined to the generated text, and surface explanations are not always causally responsible for predictions~\citep{turpinLanguageModelsDont2023}. The second strand applies latent probes to LLM activations, but typically uses them as static handles for monitoring or steering at a fixed point in the response~\citep{chenPersonaVectorsMonitoring2025}. Our proposed polylogue perspective connects the two: it inherits the functional decomposition of the first and the linear-direction methods of the second, and adds a temporal axis that tracks \emph{which} latent traits engage \emph{when}. We then ask whether the resulting dynamics are predictive of, and actionable for, model behavior.

To test this perspective, we introduce a mechanistic pipeline for open-weight reasoning models that constructs persona vectors via contrastive activation differences, tracks their stepwise alignment with model activations during CoT, and quantifies multi-persona dynamics across tasks and model families. Our contributions are as follows:
\paragraph{C1 (Polylogue Faithfulness Analysis).} We evaluate the polylogue along three axes. \emph{Semantic faithfulness} asks whether persona activations align with the persona expressed in the surface text. \emph{Functional faithfulness} asks whether polylogue features predict correctness. \emph{Causal faithfulness} asks whether these features are actionable as steering targets.
\paragraph{C2 (Stress Test of Paragraph-Conditioned Steering).} We convert predictive persona-position features into paragraph-conditioned activation interventions. The resulting steering improves accuracy on three models but substantially degrades one, revealing both the potential and the current instability of stage-aware persona steering.

\section{Persona Vectors}

\begin{figure*}[t]
    \centering
    \includegraphics[width=\textwidth]{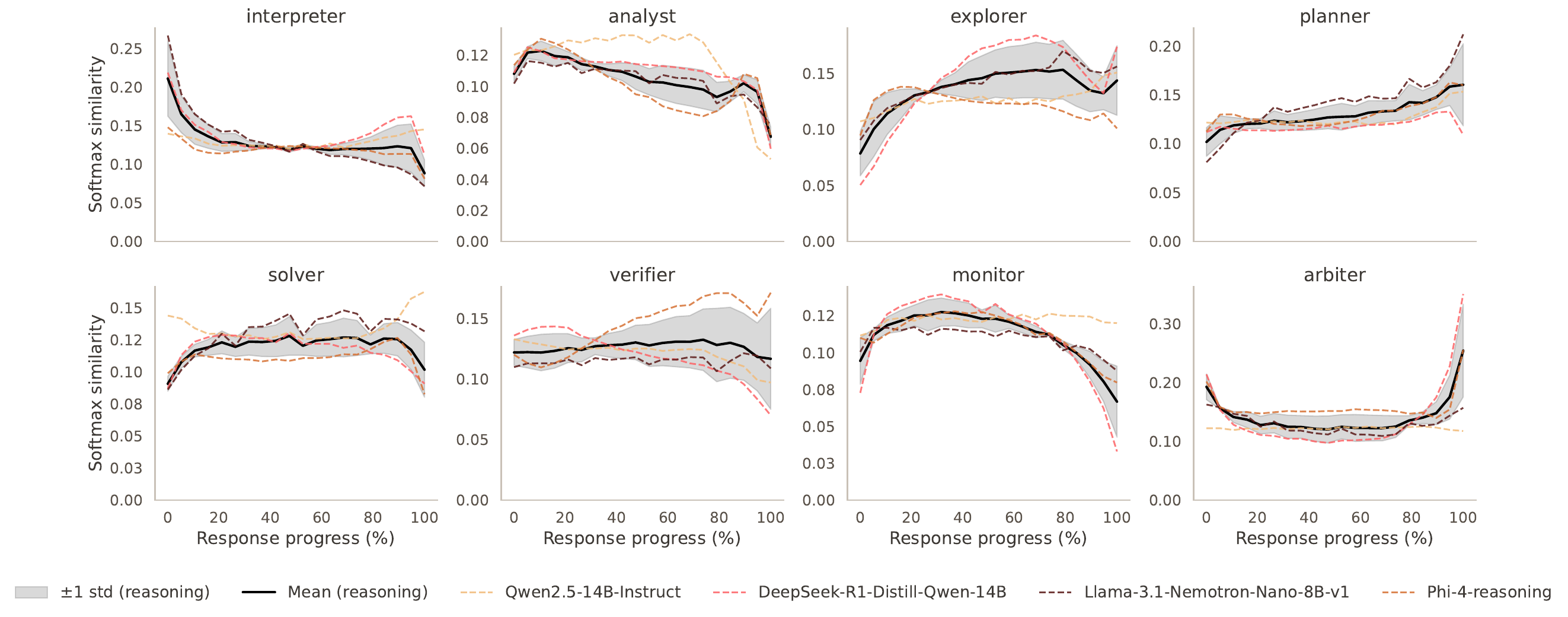}
    \caption{Latent trait alignment over response progress. Softmax-normalized mean similarity $\bar{s}_k(t)$ of persona traits as a function of normalized response position.}
    \label{fig:tag_timeline_similarities}
\end{figure*}

\subsection{Deriving the Set of Personas}
Cognitive psychology has established theories on how to segment problem solving into a sequence of functional steps, such as first understanding the task, then planning an approach, executing it, and verifying intermediate results. Here, we adopt the Episode Theory by \citet{schoenfeldMathematicalProblemSolving1985} to formalize this idea and decompose the generated CoT traces of LLMs into functional episodes in latent space, following prior work \citep{liSchoenfeldsAnatomyMathematical2025,liUnderstandingThinkingProcess2025}.
Concretely, we map \citeauthor{schoenfeldMathematicalProblemSolving1985}'s eight reasoning episodes as operationalized in \citet{liSchoenfeldsAnatomyMathematical2025,liUnderstandingThinkingProcess2025} to eight reasoning personas: \emph{Interpreter} (Read), \emph{Analyst} (Analyze), \emph{Planner} (Plan), \emph{Solver} (Implement), \emph{Explorer} (Explore), \emph{Verifier} (Verify), \emph{Monitor} (Monitor), and \emph{Arbiter} (Answer). Full descriptions are provided in Table~\ref{tab:reasoning_personas}.

\begin{table}[ht]
\footnotesize
\centering
\small
\begin{tabularx}{\linewidth}{l l X}
\toprule
\textbf{Persona} & \textbf{Episode} & \textbf{Description} \\
\midrule
Interpreter & Read      & Focuses on understanding the problem statement by parsing and restating it. \\
Analyst     & Analyse   & Identifies underlying structure, constraints, and relevant concepts to clarify what must be solved. \\
Planner     & Plan      & Thinks ahead strategically, outlines structured approaches, and acts according to a clear, organised plan. \\
Solver      & Implement & Executes a chosen strategy through explicit calculations or logical steps. \\
Explorer    & Explore   & Generates and tests alternative ideas or hypotheses to search for promising solution paths. \\
Verifier    & Verify    & Checks intermediate results and final conclusions for correctness and consistency. \\
Monitor     & Monitor   & Regulates the reasoning process by tracking progress, detecting confusion, and adjusting direction when needed. \\
Arbiter     & Answer    & Commits to a final answer and presents it clearly and decisively. \\
\bottomrule
\end{tabularx}
\caption{The eight reasoning personas, each mapped to a Schoenfeld episode \citep{schoenfeldMathematicalProblemSolving1985}.}
\label{tab:reasoning_personas}
\end{table}

\subsection{Extracting and Steering Persona Vectors}

\begin{figure}[h]
    \centering
    \includegraphics[width=\linewidth]{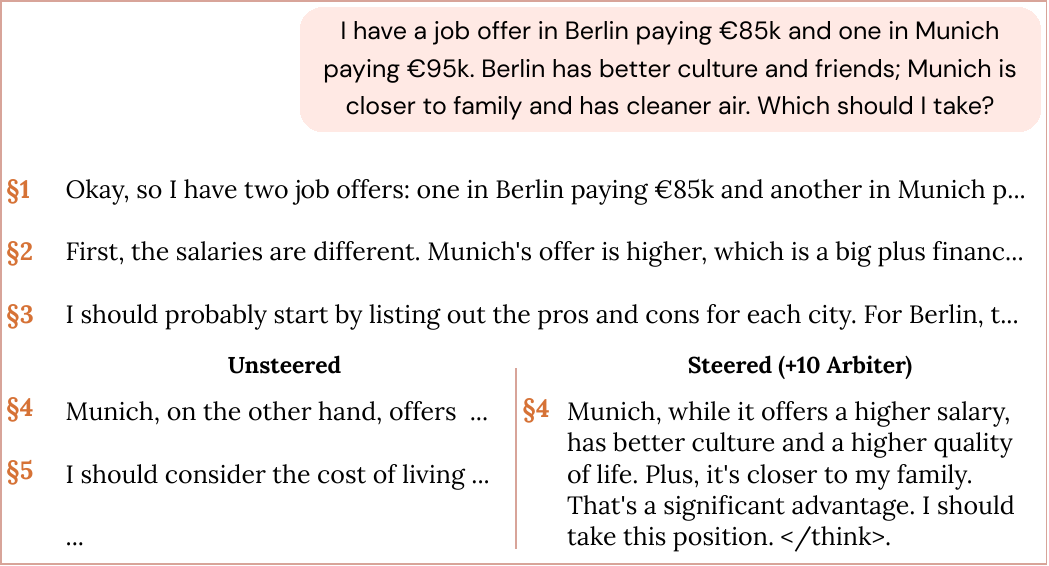}
    \caption{Example generation with and without steering using DeepSeek-R1-Distill-Qwen-14B and the arbiter persona vector. Steering towards the arbiter persona in paragraph 4 makes the reasoning stop after committing to a solution.}
    \label{fig:steering}
\end{figure}

To generate the persona vectors, we adopt the extraction pipeline proposed in \citet{chenPersonaVectorsMonitoring2025} which uses the \emph{difference in means} method \citep{subramaniExtractingLatentSteering2022, rimskySteeringLlama22024, turnerSteeringLanguageModels2024}. For each persona, we construct a trait-inducing system prompt (e.g., ``Respond like a strategic planner.'') and a trait-inhibiting system prompt (e.g., ``Do not plan or outline.''), generate responses under each condition, and define the persona vector at layer $l$ as the difference of mean residual stream activations: \[v_l = \overline{a_l}(Y^+) - \overline{a_l}(Y^-),\] where $Y^+$ and $Y^-$ are response sets under the inducing and inhibiting prompts respectively, and $\overline{a_l}(\cdot)$ is the mean activation over response tokens (restricted to post-\texttt{</think>} tokens when a reasoning trace is present).

After obtaining persona vectors, we can steer activations to induce a certain persona as seen in Figure~\ref{fig:steering}. Concretely, we add the persona vector to the residual stream at each generation step $t$: $\tilde{a}_l(t) = a_l(t) + \alpha v_l$,
where $\alpha$ determines the strength of the intervention. We empirically select the best layer and coefficient per model, see Appendix~\ref{app:selecting}.

\subsection{Monitoring Active Personas during Reasoning}
\label{sec:monitoring}

We monitor persona expression by projecting model activations onto persona vectors. Instead of computing a single alignment score at a fixed token (e.g., the final prompt token), we apply this projection at every generation step, capturing how episode engagement evolves over the course of CoT generation. For persona $k$ at step $t$, the alignment score is 
\[s_{k,t} = \frac{\langle v_k, a_t \rangle}{\|v_k\|}. \]
This yields a time series $\{s_{k,t}\}_{t=1}^{T}$ per persona. Stacking across all $K$ personas gives a multivariate time series that we call the \emph{polylogue}, which represents dynamics among latent persona directions during reasoning.

Figure~\ref{fig:tag_timeline_similarities} illustrates this view averaged across responses: the \emph{interpreter} dominates the opening of the trace and the \emph{arbiter} the close, while \emph{explorer} and \emph{solver} remain consistently engaged throughout the middle. This pattern, distinct personas active at distinct phases of generation, is the kind of latent structure the polylogue aims to reveal.

Beyond mean alignment, we summarize each polylogue with a compact set of descriptors. For each persona $k$ we compute three trait-level quantities: the \emph{average alignment} $\bar{s}_k$ over the trace (overall engagement strength), the \emph{volatility} (standard deviation of $s_{k,t}$, indicating whether the trait is stably or only intermittently engaged), and the \emph{final-step similarity} $s_{k,T}$ (engagement at the point of answer commitment). Two further descriptors capture the multivariate structure: letting $k^*(t) = \arg\max_k s_{k,t}$ be the dominant persona at step $t$, the \emph{dominance entropy} (normalized entropy of $k^*$ over the trace) measures whether reasoning is governed by a few directions or spread across many, and the \emph{switching rate} (fraction of steps where $k^*(t) \neq k^*(t-1)$) measures how often the leading persona turns over.

\begin{figure*}[ht]
    \centering
    \includegraphics[width=\textwidth]{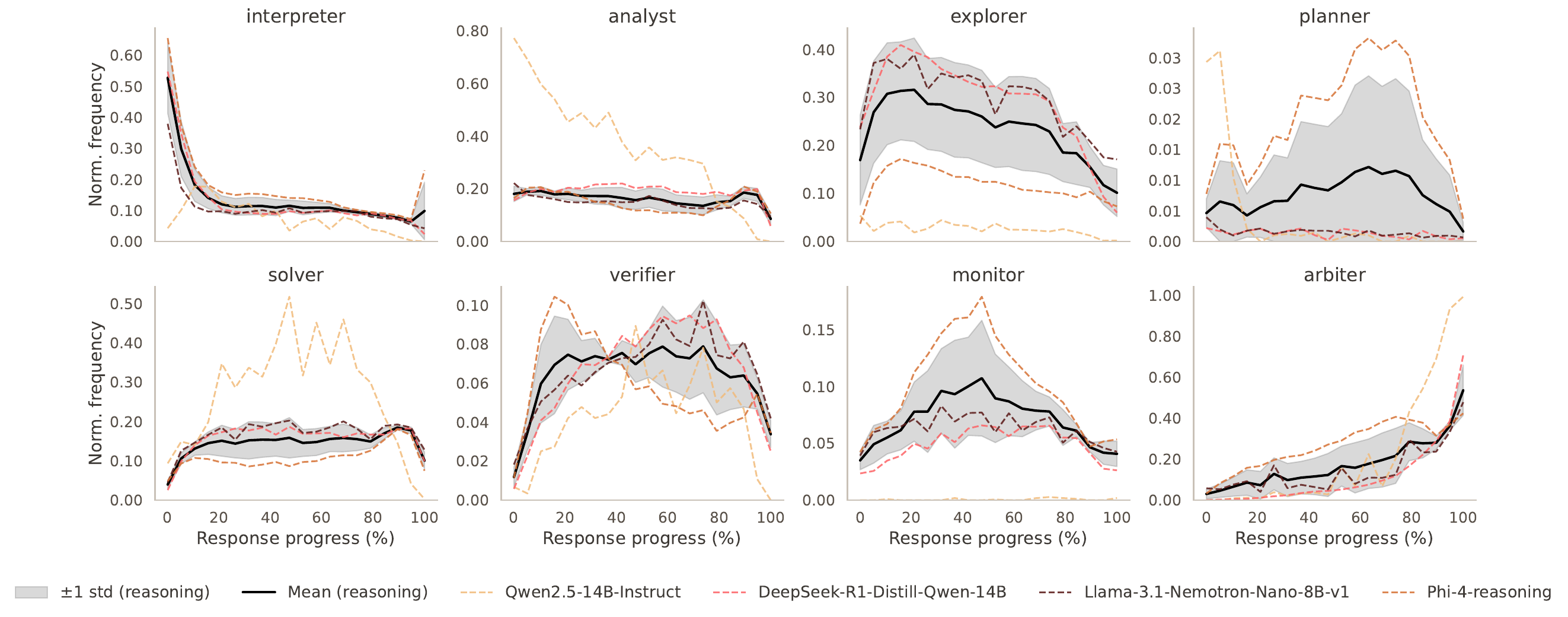}
    \caption{Paragraph label distribution over response progress. Fraction of paragraphs assigned each persona label at each normalized progress bin. The \emph{interpreter} dominates the opening bins and is almost absent by the end, while the \emph{arbiter} shows the complementary pattern and consolidates the final paragraphs. In the middle of the trace, the \emph{explorer} persona is essentially absent for the non-reasoning Qwen2.5-14B-Instruct, separating it from the three reasoning models.}
    \label{fig:tag_timeline_labels}
\end{figure*}

\subsection{Persona Vector Space Analysis}
\label{sec:persona_space}

Having fixed a set of eight reasoning personas, we ask whether this choice is well justified by analyzing their \textit{diversity} and \textit{compactness}. While diversity asks whether the personas capture genuinely different directions, compactness measures how many independent directions the space actually spans. To characterize the selection, we analyze the \emph{latent trait space} by stacking the $K$ vectors as rows of a matrix $V \in \mathbb{R}^{K \times d}$. 

We quantify diversity as the average pairwise cosine distance $1 - |\cos(\mathbf{v}_i, \mathbf{v}_j)|$ between persona vectors, which measures angular separation while ignoring sign, so higher values indicate more distinct directions. We quantify compactness by the \emph{effective rank} of $V$, the participation ratio of its singular-value spectrum. It reveals whether the persona vectors span many independent directions or instead collapse onto a lower-dimensional subspace. Concretely, it counts how many directions carry comparable variance: it equals $K$ when all directions matter equally and approaches $1$ when a single direction dominates. Formal definitions are given in Appendix~\ref{app:persona_space}.

\begin{table}[htbp]
\footnotesize
\centering
\begin{tabularx}{\linewidth}{Xrr}
\toprule
Model & Dist & Rank \\
\midrule
Qwen2.5-14B-Instruct & 0.73 & 3.74 \\
DeepSeek-R1-Distill-Qwen-14B & 0.52 & 2.83 \\
Phi-4-reasoning & 0.58 & 2.68 \\
Llama-3.1-Nemotron-Nano-8B-v1 & 0.45 & 1.85 \\
\bottomrule
\end{tabularx}
\caption{Persona vector space geometry per model. \emph{Dist}: average pairwise cosine distance $1 - |\cos(\mathbf{v}_i, \mathbf{v}_j)|$ over all trait-vector pairs. \emph{Rank}: effective rank, the participation ratio of the SVD spectrum. Higher values mean more distinct and more independent directions. All effective ranks fall well below $K=8$, indicating that the persona directions partly overlap and likely span a lower-dimensional subspace.}
\label{tab:persona_space_geometry}
\end{table}

\paragraph{Results.} Table~\ref{tab:persona_space_geometry} reports both quantities per model, Figure~\ref{fig:avg_cosine_distance_heatmap} (Appendix~\ref{app:persona_space}) breaks the distances down by persona pair. Pairwise distances are moderate (0.45--0.73), with some near-collinear pairs such as \emph{solver}--\emph{verifier} ($0.25$) and some near-orthogonal pairs such as \emph{interpreter}--\emph{planner} ($0.82$), indicating a relatively diverse set of directions. At the same time, the effective ranks (1.85--3.74) fall well below $K=8$, showing that the extracted persona vectors occupy a compact, lower-dimensional subspace.

\section{Faithfulness Experiments}

All faithfulness experiments use a subset of MMLU-Pro~\citep{wangMMLUProMoreRobust2024} on four open-weight models. MMLU-Pro is a diverse multiple-choice benchmark covering 14 domains, designed to measure broad capabilities of the models. For details, see Appendix \ref{app:mmlu_pro}.

\subsection{Semantic Faithfulness}
\label{sec:semantic_faithfulness}

\begin{figure*}[ht!]
    \centering
    \includegraphics[width=\textwidth]{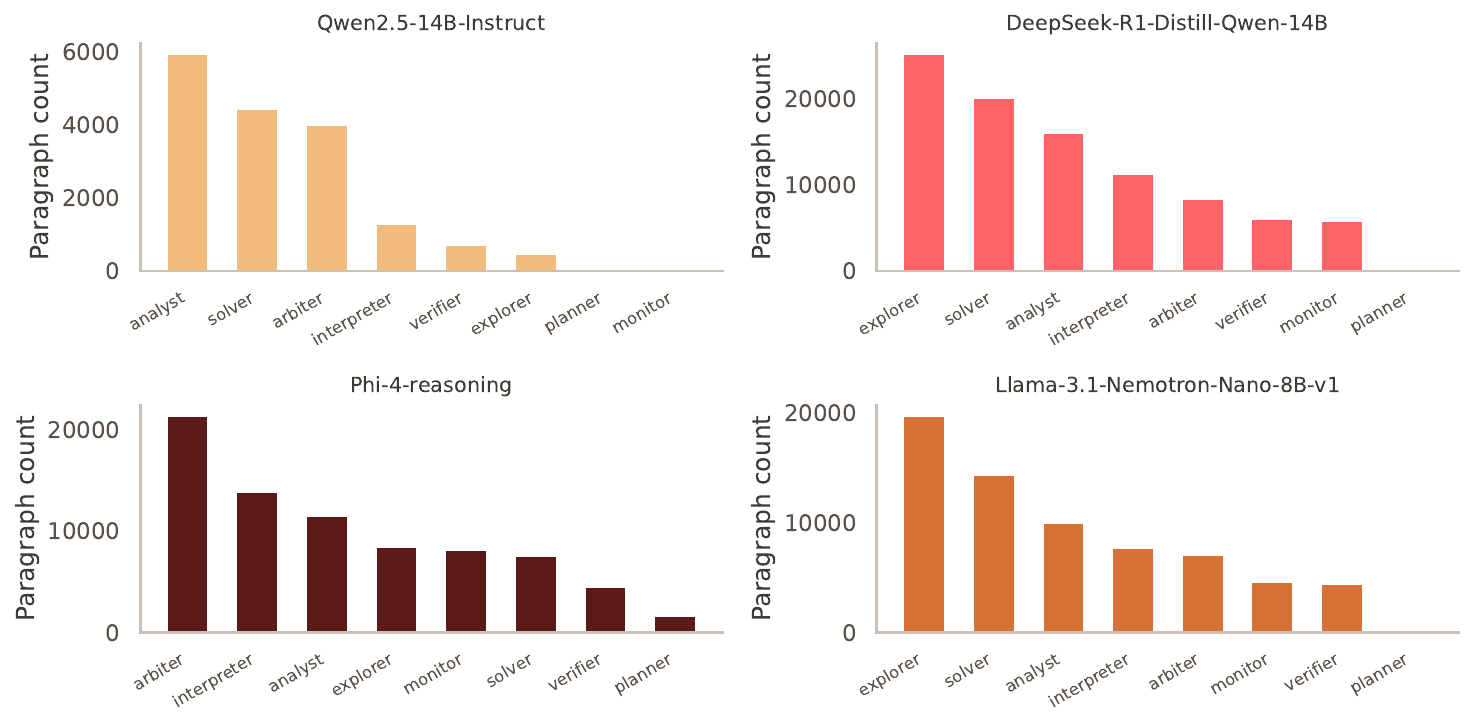}
    \caption{Paragraph label distribution across models computed on MMLU-Pro responses.}
    \label{fig:label_distribution}
\end{figure*}

\paragraph{Surface vs.\ latent persona manifestation.}
First, we analyze whether the persona expressed in the generated text aligns with the persona most salient in the model's activations at the same reasoning step. To assess this, we compare two views of the same CoT: a \emph{surface} view, derived from the text, and a \emph{latent} view, derived from the activations.

To construct the surface view, we segment each CoT into paragraphs and assign a persona label to every paragraph using Llama-3.3-70B-Instruct. Figure~\ref{fig:tag_timeline_labels} shows the frequency of labels across responses for different relative time steps. A consistent structural template emerges across models: the \emph{interpreter} dominates the opening, where the problem is parsed, and is almost absent by the end, while the \emph{arbiter} shows the complementary pattern, consolidating the final paragraphs where the model commits to an answer; \emph{explorer} and \emph{solver} dominate the core reasoning in between. The middle of the trace separates the three reasoning models from the non-reasoning baseline (Qwen2.5-14B-Instruct): the \emph{explorer} persona is essentially absent in the instruction-tuned model, which moves directly from interpretation to commitment, and the \emph{analyst} persona, sustained throughout reasoning in the three reasoning models, decays over the trace in the instruction-tuned baseline.

The latent timeline of Figure~\ref{fig:tag_timeline_similarities} broadly mirrors this template but is markedly smoother, and the reasoning-vs.-instruction-tuned contrast is less pronounced there: the \emph{explorer} and \emph{analyst} directions stay engaged across all four models, even when the instruction-tuned model does not verbalize the corresponding behavior. This discrepancy may indicate latent activity not expressed in the text, but it may also reflect limited semantic specificity of the extracted directions.

To test this correspondence quantitatively, we formulate semantic faithfulness as a ranking problem. For each paragraph, we compute the mean projection of hidden activations onto each persona vector across the paragraph's tokens,\footnote{Activation projections are whitened with the Mahalanobis transform. See Appendix~\ref{app:mahalanobis}.} then rank the eight personas by these scores. We quantify alignment using Mean Reciprocal Rank (MRR).

We compare the polylogue ranking against two baselines: a uniform random baseline and a frequency baseline that ranks personas by their empirical label frequency. See Appendix~\ref{app:mrr} for details and the formal definition.

\begin{table}[ht]
\centering
\footnotesize
\begin{tabularx}{\linewidth}{Xrrrr}
\toprule
Model & Rnd & Frq & Poly \\
\midrule
Qwen2.5-14B-Instruct & 0.34 & \textbf{0.59} & 0.35 \\
DeepSeek-R1-Distill-Qwen-14B & 0.34 & 0.50 & \textbf{0.51}\\
Phi-4-reasoning & 0.34 & 0.49 & \textbf{0.54} \\
Llama-3.1-Nemotron-Nano-8B-v1 & 0.34 & \textbf{0.51} & 0.47 \\
\bottomrule
\end{tabularx}
\caption{Mean reciprocal rank (MRR) of activation similarities to LLM-assigned paragraph labels. Rnd: uniform random baseline; Frq: frequency baseline (traits ranked by empirical label frequency); Poly: Polylogue. Bold: best result per row.}
\label{tab:mrr}
\end{table}

\begin{table*}[ht]
\footnotesize
\centering
\setlength{\tabcolsep}{4pt}
\begin{tabular}{lrrrrrrr}
\toprule
Model & \% Correct & \multicolumn{2}{c}{Random} & \multicolumn{2}{c}{Polylogue} & \multicolumn{2}{c}{Activation Baseline} \\
\cmidrule(lr){3-4} \cmidrule(lr){5-6} \cmidrule(lr){7-8}
 & & Acc $\pm$ std & AUC $\pm$ std & Acc $\pm$ std & AUC $\pm$ std & Acc $\pm$ std & AUC $\pm$ std \\
\midrule
Qwen2.5-14B-Instruct & 62.1\% & 0.61 $\pm$ 0.01 & \textbf{0.59 $\pm$ 0.03} & 0.61 $\pm$ 0.01 &\textbf{ 0.59 $\pm$ 0.02} & \textbf{0.63 $\pm$ 0.00} & 0.51 $\pm$ 0.01 \\
DeepSeek-R1-Distill... & 60.3\% & 0.75 $\pm$ 0.02 & 0.78 $\pm$ 0.02 & 0.75 $\pm$ 0.01 & 0.81 $\pm$ 0.00 & \textbf{0.78 $\pm$ 0.02} & \textbf{0.84 $\pm$ 0.01} \\
Phi-4-reasoning & 56.0\% & 0.80 $\pm$ 0.01 & 0.84 $\pm$ 0.01 & 0.81 $\pm$ 0.02 & \textbf{0.87 $\pm$ 0.01} & \textbf{0.82 $\pm$ 0.02} & 0.85 $\pm$ 0.02 \\
Llama-3.1-Nemotron... & 42.0\% & 0.68 $\pm$ 0.02 & 0.70 $\pm$ 0.01 & \textbf{0.73 $\pm$ 0.03} & \textbf{0.76 $\pm$ 0.03} & 0.71 $\pm$ 0.01 & \textbf{0.76 $\pm$ 0.01} \\
\bottomrule
\end{tabular}
\caption{Correctness prediction performance per model using L1 logistic regression. Activation Baseline uses last-token hidden states (PCA-reduced); Random uses random-vector projections; Polylogue uses persona-vector features.}
\label{tab:correctness_performance}
\end{table*}

\paragraph{Results.} Table~\ref{tab:mrr} reports MRR for the four models. Polylogue exceeds the random-ranking baseline on all models, indicating above-chance correspondence between latent rankings and textual labels. However, it beats the frequency baseline on only two of four models, so surface--latent agreement is weak overall. Two factors plausibly contribute. First, the label distribution is heavily skewed (Figure~\ref{fig:label_distribution}), which makes the frequency baseline strong by construction. Second, reasoning modes are not mutually exclusive: a single paragraph can involve exploration, solving, and verification at once, so a single label understates latent agreement. Both remain hypotheses; on the current evidence, we regard semantic faithfulness as only partially supported, consistent with findings that surface-level CoT does not always mirror internal computation \citep{turpinLanguageModelsDont2023}.

\subsection{Functional Faithfulness}
\label{sec:functional_faithfulness}

\paragraph{Method and Baselines.} To assess the usefulness of polylogue features, we investigate whether they are predictive of correctness. We fit an L1-regularized (Lasso) logistic regression (details in Appendix~\ref{app:functional_faithfulness}). Through the resulting sparse feature selection, we identify \emph{which} persona signals matter and \emph{in which paragraph}, yielding an interpretable set of predictive features. Performance is evaluated using 5-fold cross-validation, reporting accuracy and AUC-ROC.

We compare polylogue against two non-interpretable references with equal or greater representational capacity. The \emph{random-projection baseline} replaces persona vectors with random unit vectors, which computationally resembles polylogue features, but using semantically meaningless personas. This isolates the contribution of interpretable directions over general activation geometry. The \emph{activation baseline} uses the last-token hidden state of the response, reduced via PCA before classifier fitting. Since this baseline has access to the entire generation state, it serves as a reference for what is linearly extractable from activations.

\paragraph{Results and Feature Analysis.} Table~\ref{tab:correctness_performance} reports how well we can predict correctness across the four models. Polylogue features approach the activation baseline, and do not clearly outperform random projections. The predictive signal can therefore not be attributed to persona semantics: much of it is recoverable from generic low-dimensional projections of the activation geometry. This is consistent with the low effective rank of the persona set (Section~\ref{sec:persona_space}) --- the eight directions span only a few independent dimensions, so a comparably sized set of random directions covers similar variance. What the persona directions add is not accuracy but interpretability: the fitted coefficients can be read as persona-and-position statements, which we exploit next.

The sparsity of the L1 logistic regression fit lets us go beyond predictive accuracy and read the coefficients to derive insights regarding which latent characteriztics are associated with correct versus incorrect answers. 

\begin{table}[ht]
\footnotesize
\centering
\begin{tabular}{clc}
\toprule
Rank & Feature & Coef. \\
\midrule
1 & final sim monitor & -0.74 \\
2 & para 0 interpreter & +0.16 \\
3 & para 7 interpreter & -0.12 \\
4 & para 19 planner & +0.12 \\
5 & para 18 arbiter & +0.12 \\
\bottomrule
\end{tabular}
\caption{Top coefficients of the L1 logistic regression on polylogue features for DeepSeek-R1-Distill-Qwen-14B. ``para $p$ \textit{persona}'' denotes the mean alignment of \textit{persona} over paragraph $p$; ``final sim \textit{persona}'' is the alignment at the final generation step.}
\label{tab:correctness_coefficients_deepseek}
\end{table}

Table~\ref{tab:correctness_coefficients_deepseek} shows the top coefficients for DeepSeek-R1-Distill-Qwen-14B (coefficients for other models are in Appendix~\ref{app:all_coefficients}). The strongest predictor is \emph{monitor} activation at the final step, with a large negative weight ($-0.74$): when the model is still monitoring its own reasoning at the end of the trace, the answer tends to be wrong. Smaller positive weights relate to \emph{interpreter} activity in the opening paragraph and \emph{planner} and \emph{arbiter} activity near answer commitment, which suggests that correct responses follow a clean arc from problem comprehension to the final committed answer.

\subsection{Causal Faithfulness}
\label{sec:causal_faithfulness}

\paragraph{Steering reasoning personas.}
Next, we ask whether the polylogue is \emph{useful} in addition to being indicative of reasoning traits. That is, whether the patterns surfaced by the functional analysis can be turned into an intervention that improves model behavior.

We translate the \textit{paragraph features} (e.g., ``para 0 interpreter'') into a steering strategy. Specifically, we select the top-5 features by absolute coefficient magnitude and steer the model towards the corresponding persona in the corresponding paragraph. Features with positive coefficients are added; those with negative coefficients are subtracted. We compare steered generations against the unsteered baseline on the same prompts. For more details, see Appendix~\ref{app:causal_faithfulness}.

\begin{table}[h]
\footnotesize
\centering
\begin{tabularx}{\linewidth}{Xrrrr}
\toprule
Model & \multicolumn{2}{c}{Acc (\%)} & \multicolumn{2}{c}{F1 (\%)} \\
\cmidrule(lr){2-3}\cmidrule(lr){4-5}
 & Base & Steered & Base & Steered \\
\midrule
Qwen2.5-14B-Instruct & 63.3 & \textbf{64.1} & 63.3 & \textbf{64.3} \\
DeepSeek-R1-Distill... & 62.3 & \textbf{64.5} & 62.8 & \textbf{65.0} \\
Phi-4-reasoning & \textbf{53.0} & 38.5 & \textbf{54.3} & 39.8 \\
Llama-3.1-Nemotron... & 44.4 & \textbf{48.8} & 44.4 & \textbf{48.9} \\
\bottomrule
\end{tabularx}
\caption{MMLU-Pro: steered vs.\ baseline (vLLM). Bold indicates the better value per metric.}
\label{tab:mmlu_pro_steering}
\end{table}

\paragraph{Results.} 
Table~\ref{tab:mmlu_pro_steering} reports steered vs.\ baseline accuracy and F1 on MMLU-Pro. Steering improves accuracy on three of the four models. Phi-4-reasoning is the exception: steering substantially degrades its performance, showing that the intervention does not transfer reliably across models. We discuss concrete steps towards a more robust intervention in Section~\ref{sec:discussion}.

\section{Discussion and Future Work}
\label{sec:discussion}

\paragraph{Persona Set.} This work uses eight reasoning personas grounded in Schoenfeld's Episode Theory. The resulting vectors were moderately diverse yet compact, with an effective rank well below eight. The extracted vectors form a low-dimensional space, although it remains unclear whether this reflects the structure of reasoning personas, semantic overlap between the chosen roles, or shared artifacts of the extraction prompts. The compactness may also partly stem from extraction, since the difference-in-means vectors are all built from the same prompt template, which can lower the effective rank on its own. A natural next step is to construct a broader and more diverse persona inventory and revisit diversity and compactness under that wider lens. Such an analysis should also study the space \emph{after} projection onto activations, not just the static vectors, which capture only part of how the directions interact with model state.

\paragraph{Monitoring.} Polylogue features predict correctness on par with activation baselines, but the relationship is correlational: the polylogue may reflect a reasoning process rather than drive it. Our characterization of the polylogue was also relatively simple, consisting of per-paragraph statistics, volatility, and a handful of cross-persona summaries. A deeper analysis of the latent dynamics would likely surface richer structure than the descriptors we used. This could entail characterizing the polylogue as a latent graph over personas, with edges capturing transitions.

\paragraph{Steering.} The polylogue also points to concrete steering targets: which persona direction to amplify or suppress, and in which region of the reasoning trace. The intervention we built from this is intentionally simple, with a small set of predictive features deciding what to steer in each paragraph. It improves correctness on three of the four models. We read this as evidence that stage-aware steering is possible, but that fixed paragraph-level rules are too rigid for robust control. For Phi-4-reasoning, where steering degraded performance, two checks could sharpen the picture: a careful inspection of steered outputs to find the source of error, distinguishing fluency degradation from mis-timed persona activation, and a re-selection of the steering layer and coefficient on the final persona set.

A natural next step is adaptive steering: instead of fixed paragraph-level rules, a controller could decide online what to steer, when to steer, and with what strength, conditioned on the unfolding reasoning content.

\paragraph{Analysis axes.} Our evaluation analyzed only one behavioral axis: correctness. This approach extends naturally to other axes of behavior, such as alignment and diversity. Future work could introduce persona inventories tailored to those axes rather than borrowed from the reasoning regime.

\section{Conclusion}

We investigated whether latent persona directions can be used to characterize LLM reasoning. To this end, we introduced the \emph{polylogue}: the time-varying pattern of alignments between persona vectors and hidden activations over a chain-of-thought. We studied it along three notions of faithfulness --- semantic, functional, and causal --- on correctness with MMLU-Pro.

Our findings are encouraging but partial. Polylogue features predict correctness competitively with strong activation baselines while using only eight interpretable directions. Acting on that signal is harder: a simple paragraph-conditioned steering scheme derived from the same features improves correctness on three of four models, but its mixed performance indicates that fixed paragraph-level rules are not sufficient for robust control. Taken together, these results position dynamic persona-direction analysis as a promising lens on reasoning-time model behavior, while leaving open which persona inventories, controllers, and validation procedures are needed to make it actionable.

\section{Limitations}
This work has several limitations. First, we rely on an LLM judge in two places in the pipeline: when scoring trait expression to calibrate the steering coefficient and layer (Appendix~\ref{app:selecting}), and when tagging paragraphs with persona labels in the semantic-faithfulness analysis (Section~\ref{sec:semantic_faithfulness}). Both steps should be validated against human annotators. Second, the experimental setting is relatively synthetic: the benchmark consists of curated multiple-choice tasks rather than open-ended use. Third, all experiments are conducted in English, and the results would have to be validated in additional languages before any claim of cross-lingual generality. Fourth, the layer and coefficient used for steering were selected on a set of probe personas that differs from the eight reasoning personas used in the rest of the paper, a legacy of an earlier exploratory phase (Appendix~\ref{app:selecting}); re-running the selection on the final persona set could yield a better configuration. Fifth, since generation under activation steering is computationally expensive, the causal evaluation is limited to 504 questions per model, and we report no significance estimates; the smaller steering gains should therefore be interpreted with caution.

\section{Ethical Considerations}

This work develops methods for monitoring and intervening in the latent dynamics of LLMs. While these methods may support more interpretable and reliable behavior, they also introduce risks.

\paragraph{Analysis Dimensions.} Our intervention is optimized only for correctness on MMLU-Pro. However, correctness does not ensure that a response is safe, fair, well calibrated, or aligned with user values. Optimizing for this single objective may therefore create unintended trade-offs. For example, suppressing the \emph{monitor} persona could be harmful in high-stakes settings, where uncertainty and self-checking may be important. Before deployment, steering strategies should be evaluated along multiple dimensions, including alignment and diversity.

\paragraph{Automation.} Extracting and evaluating persona vectors requires many model generations (2,000 for each vector) and relies on automated LLM judgments. This automation makes the pipeline scalable, but makes human review more costly. It may also reproduce biases or errors from the extraction prompts, benchmarks, or judge models. Human validation and oversight therefore remain important, especially when evaluating dimensions such as alignment and diversity.

\paragraph{Anthropomorphism.} We use the terms \emph{persona} and \emph{polylogue} as functional descriptions of activation patterns. They should not be understood as evidence that language models possess distinct identities, subjective experiences, or consciousness. Nevertheless, this terminology may unintentionally contribute to the anthropomorphization of language models and should therefore be interpreted with care.

\section*{Acknowledgments}
We thank Alireza Modirshanechi for the insightful discussions. The authors gratefully acknowledge the scientific support and resources of the AI service infrastructure LRZ AI Systems provided by the Leibniz Supercomputing Centre (LRZ) of the Bavarian Academy of Sciences and Humanities (BAdW), funded by Bayerisches Staatsministerium für Wissenschaft und Kunst (StMWK). We also acknowledge the use of the HPC cluster at Helmholtz Munich for the computational resources used in this study.

\bibliography{custom}

\appendix

\section{Selecting layers and coefficients}
\label{app:selecting}

We follow the empirical selection procedure of \citet{chenPersonaVectorsMonitoring2025}. For each latent vector \( v_l \), we evaluate different combinations of layer \( l \) and steering coefficient \( \alpha \) using full activation steering during generation.

For each configuration, we generate responses and assess two characteriztics using an LLM judge (Llama-3.3-70B-Instruct): (i) \emph{trait expression}, measuring the degree to which the target behavior is present, and (ii) \emph{coherence}, measuring the fluency and consistency of the response. The LLM judge is tasked to score each characteriztic between $0$ and $100$.

To obtain a scalar score, we aggregate over the model’s output distribution rather than relying on a single token. Concretely, we extract the logits $\ell_k$ corresponding to numeric tokens ($0$--$100$), convert them into (unnormalized) probabilities $\exp(\ell_k)$, and compute the expected value:
\[
\text{score} = \frac{\sum_{k=0}^{100} k \cdot \exp(\ell_k)}{\sum_{k=0}^{100} \exp(\ell_k)}.
\]
This yields a probability-weighted average score, capturing uncertainty in the model’s prediction. The normalization in the denominator ensures that the score corresponds to the expectation under the model’s distribution restricted to numeric tokens. If the total probability mass assigned to valid numeric tokens is too low, the score is discarded.

To select the optimal configuration, we define an objective that captures the trade-off between trait control and generation quality, given by the weighted geometric mean
\[
\mathcal{O}(l, \alpha) \;=\; \text{score}^{\,\beta} \cdot \text{coherence}^{\,1-\beta},
\qquad \beta \in (0,1).
\]
The multiplicative form ensures that a near-zero value in either factor drives the objective to zero, preventing one factor from compensating for the other (e.g., high trait expression in degenerate outputs). The exponent $\beta$ controls the trade-off: $\beta = 0.5$ recovers the unweighted geometric mean, while $\beta > 0.5$ favors trait expression.

We set $\beta = 0.7$, reflecting the asymmetry of the selection problem: coherence remains in a narrow high range across viable configurations, whereas trait expression varies substantially with $(l, \alpha)$ and is the quantity we primarily wish to maximize. We compute $\mathcal{O}(l, \alpha)$ for all evaluated pairs, average over evaluation prompts, and select the layer and coefficient with the highest mean objective.

Table~\ref{tab:persona_config} shows the optimal layer and coefficient selected for each model. Since evaluating responses across all combinations of personas, layers, and coefficients is computationally expensive, we restricted the search to a subset: four personas (\emph{solver}, \emph{ethical}, \emph{creativity}, \emph{agreeableness}), five layers uniformly spaced across the model, and five coefficients (0.5, 1.0, 1.5, 2.0, 2.5). The number of traits reported in the table (34--35) and the personas used for selection do not match the eight reasoning personas used elsewhere in our analysis: this configuration was tuned during an earlier exploratory phase in which we constructed a larger candidate set of personas, and we reuse the resulting layer and coefficient choices here.

\begin{table}[h]
\footnotesize
\centering
\begin{tabular}{lrrr}
\toprule
Model & Num traits & Layer & Coef \\
\midrule
Qwen2.5-14B-Instruct & 35 & 0 & 2.0 \\
DeepSeek-R1-Distill... & 34 & 23 & 2.0 \\
Phi-4-reasoning & 34 & 9 & 2.5 \\
Llama-3.1-Nemotron... & 35 & 31 & 0.5 \\
\bottomrule
\end{tabular}
\caption{Persona vector configuration per model.}
\label{tab:persona_config}
\end{table}

\section{Persona Vector Space Analysis}
\label{app:persona_space}

\begin{figure*}[ht]
    \centering
    \includegraphics[width=\textwidth]{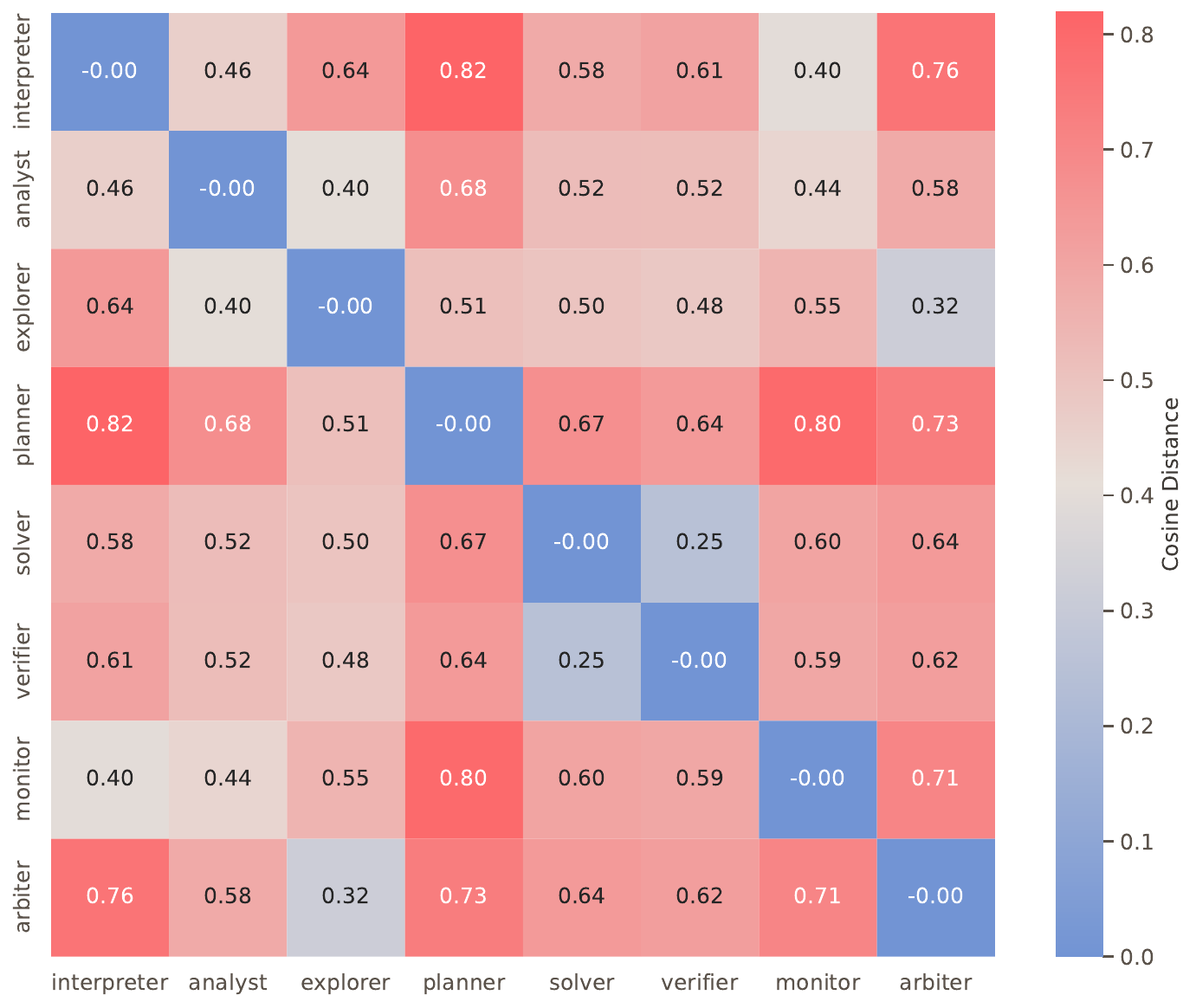}
    \caption{Average pairwise cosine distance between persona vectors, averaged across all models. Each cell $(i, j)$ reports $1 - |\cos(\mathbf{v}_i, \mathbf{v}_j)|$, so a value of 0 indicates collinear vectors (parallel or anti-parallel) and a value of 1 indicates orthogonal vectors. Distances are overall moderate, reflecting a persona set that is neither fully collinear nor fully orthogonal.}
    \label{fig:avg_cosine_distance_heatmap}
\end{figure*}

This appendix gives the formal definitions for the persona space analysis in Section~\ref{sec:persona_space}, together with the per-pair distance heatmap (Figure~\ref{fig:avg_cosine_distance_heatmap}).

\paragraph{Persona vector diversity.} To measure diversity, we compute the average pairwise distance between persona vectors, defined as
\[
1 - \left| \cos(\mathbf{v}_i, \mathbf{v}_j) \right|.
\]
This captures angular separation while ignoring sign, and thus reflects whether traits correspond to genuinely different directions in representation space.

\paragraph{Persona vector compactness.} To assess compactness, we analyze the spectrum of the persona matrix $V$. Let $\sigma_1 \geq \sigma_2 \geq \dots \geq \sigma_K$ denote its singular values, obtained by singular value decomposition. The squared singular values $\sigma_i^2$ measure the variance carried by each orthogonal direction, and we normalize them into a distribution
\[
p_i = \frac{\sigma_i^2}{\sum_j \sigma_j^2}.
\]
The effective rank is then defined as the participation ratio
\[
r_{\text{eff}} = \frac{1}{\sum_i p_i^2} = \frac{\left(\sum_i \sigma_i^2\right)^2}{\sum_i \sigma_i^4}.
\]
Intuitively, $r_{\text{eff}}$ counts how many directions carry comparable variance: it equals $K$ when the spectrum is flat (all directions equally important) and approaches $1$ when a single direction dominates.

\section{MMLU-Pro Details}
\label{app:mmlu_pro}

We use the MMLU-Pro test set, which comprises 12{,}032 multiple-choice questions across 14 academic domains (Table~\ref{tab:mmlu_pro_categories}).

\begin{table}[ht]
\centering
\footnotesize
\begin{tabular}{lc}
\toprule
Category & $N$ \\
\midrule
Math             & 1{,}351 \\
Physics          & 1{,}299 \\
Chemistry        & 1{,}132 \\
Law              & 1{,}101 \\
Engineering      &    969 \\
Other            &    924 \\
Economics        &    844 \\
Health           &    818 \\
Psychology       &    798 \\
Business         &    789 \\
Biology          &    717 \\
Philosophy       &    499 \\
Computer Science &    410 \\
History          &    381 \\
\midrule
Total            & 12{,}032 \\
\bottomrule
\end{tabular}
\caption{MMLU-Pro test-set category distribution.}
\label{tab:mmlu_pro_categories}
\end{table}

The dataset is partitioned into two disjoint subsets, each stratified by domain, using a fixed 50/50 split per category:
\begin{itemize}
    \item \textbf{Semantic faithfulness and functional faithfulness}: 190 samples per domain ($14 \times 190 = 2{,}660$ total).
    \item \textbf{Causal faithfulness}: 36 samples per domain ($14 \times 36 = 504$ total), non-overlapping with the functional faithfulness subset.
\end{itemize}

\subsection{Generation Settings}
Responses are generated with vLLM using greedy decoding (\texttt{do\_sample=False}), a maximum sequence length of 3{,}072 tokens, and \texttt{tensor\_parallel\_size=1}. Feature extraction and classifier fitting use scikit-learn; parameters for \texttt{LogisticRegressionCV}/\texttt{LassoCV} are given in Appendix~\ref{app:functional_faithfulness}.

\subsection{Licenses}
MMLU-Pro is released under the MIT License. All models used in this work were accessed under their respective licenses: Apache 2.0 for Qwen2.5, MIT for DeepSeek-R1 and Phi-4, and the applicable NVIDIA and Llama Community licenses for Llama-3.1-Nemotron. Our released code does not redistribute model weights or the MMLU-Pro dataset. The licenses permit the research use and release of the artifacts presented in this work.

\section{Semantic Faithfulness}
\label{app:semantic_faithfulness}

\subsection{Mahalanobis Whitening}
\label{app:mahalanobis}

Persona vectors are extracted independently and need not be orthogonal: pairs of directions can be partially correlated, and per-persona projection magnitudes can differ in scale. Both effects bias a naive ranking by raw projection towards directions that simply have larger variance or that overlap with several others. To remove these confounds, we apply a global Mahalanobis whitening to projections before ranking.

Let $X \in \mathbb{R}^{N \times |\mathcal{T}|}$ stack the per-token projections $s_{k,t}$ across all responses in the dataset, with $N$ the total number of tokens. We compute the empirical mean $\mu \in \mathbb{R}^{|\mathcal{T}|}$ and a shrunk covariance
\[
\hat\Sigma \;=\; (1-\lambda)\,\Sigma \;+\; \lambda\,\bar\sigma^{2}\,I,
\]
where $\Sigma$ is the empirical covariance of the centered matrix $X-\mu$, $\bar\sigma^{2}$ is the mean of the diagonal of $\Sigma$, and $\lambda = 0.05$ provides numerical stability by pulling the covariance towards a scaled identity. The whitening matrix $W = \hat\Sigma^{-1/2}$ is obtained from the symmetric eigendecomposition $\hat\Sigma = U\Lambda U^{\top}$ as $W = U\,\Lambda^{-1/2}\,U^{\top}$, with eigenvalues clipped from below for stability. Each projection vector is then mapped to
\[
\tilde{s} \;=\; (s - \mu)\,W,
\]
yielding zero-mean, decorrelated, unit-variance scores across personas. The paragraph-mean activations used in the ranking task in Section~\ref{sec:semantic_faithfulness} are computed on these whitened scores, so that all personas contribute on a comparable scale and shared variance between directions does not double-count.

\subsection{Mean Reciprocal Rank}
\label{app:mrr}

For each paragraph $p$ in a CoT, let $\ell_p \in \mathcal{T}$ be the persona label assigned by the external LLM annotator, and let $\pi_p : \mathcal{T} \to \{1, \dots, |\mathcal{T}|\}$ be the ranking of personas by mean activation similarity over the tokens in $p$ (rank $1$ being most active). The reciprocal rank of paragraph $p$ is $1/\pi_p(\ell_p)$, and the Mean Reciprocal Rank over a set $\mathcal{P}$ of paragraphs is
\[
\mathrm{MRR} \;=\; \frac{1}{|\mathcal{P}|} \sum_{p \in \mathcal{P}} \frac{1}{\pi_p(\ell_p)}.
\]
The metric is bounded in $\bigl(0, 1\bigr]$. It equals $1$ when the text-assigned label is always the top-ranked persona, and decays geometrically as the label is pushed down the ranking ($1/2$ at rank $2$, $1/3$ at rank $3$, etc.).

\paragraph{Random baseline.}
A uniformly random ranking assigns the text label to rank $r$ with probability $1/|\mathcal{T}|$ for each $r$, yielding an expected MRR of
\[
\mathrm{MRR}_{\mathrm{rnd}} \;=\; \frac{1}{|\mathcal{T}|} \sum_{r=1}^{|\mathcal{T}|} \frac{1}{r}.
\]

\paragraph{Frequency baseline.}
The frequency baseline ranks personas once, globally, by their empirical label frequency in the dataset, with the most-frequent persona at rank $1$. Concretely, let $n_k = \sum_{p \in \mathcal{P}} \mathbb{I}[\ell_p = k]$ be the number of paragraphs labeled with persona $k$, and let $\pi^{\mathrm{frq}}$ be the ranking induced by sorting $(n_k)_{k \in \mathcal{T}}$ in decreasing order. Every paragraph is scored against this single dataset-wide ranking,
\[
\mathrm{MRR}_{\mathrm{frq}} \;=\; \frac{1}{|\mathcal{P}|} \sum_{p \in \mathcal{P}} \frac{1}{\pi^{\mathrm{frq}}(\ell_p)}
\]
so a paragraph whose label is the most common persona scores $1$, the second-most-common scores $1/2$, and so on. This is a strong baseline whenever the label distribution is skewed, since always guessing the dominant persona first is hard to beat under those conditions.

\section{Functional Faithfulness}
\label{app:functional_faithfulness}

\subsection{Experiment Details}
For each generated response, we first locate paragraph boundaries by splitting the decoded token sequence on double newlines. The feature vector consists of the full set of polylogue descriptors defined in Section~\ref{sec:monitoring}, computed at paragraph-bin granularity. Concretely, for each persona $k \in \mathcal{T}$ we extract:
\begin{itemize}
    \item Per-bin mean alignment $\bar{s}_k^{(b)}$ within each of $n_b = 20$ equal-width paragraph bins, where the paragraphs of a response are partitioned so that bin $b$ aggregates over all tokens in the paragraphs falling in the $b$-th fraction of the trace ($n_b \times |\mathcal{T}|$ features in total).
    \item The volatility of $s_{k,\cdot}$ (standard deviation over the full trace).
    \item The final-step similarity $s_{k,T}$.
\end{itemize}
We additionally include the cross-persona descriptors: the per-persona dominance share, the normalized dominance entropy, and the dominant switching rate. This yields a feature vector of dimension $n_b \cdot |\mathcal{T}| + 3 \cdot |\mathcal{T}| + 2$ per response.

\paragraph{Regression setup.}
All feature vectors are standardized (zero mean, unit variance) before fitting. For binary outcomes (correctness, alignment), we use L1-penalized logistic regression (\texttt{LogisticRegressionCV} with \texttt{solver=saga}, 10 candidate penalty values $C$, inner 5-fold cross-validation optimizing AUC-ROC). For the continuous diversity target, we use \texttt{LassoCV} with the same cross-validation scheme. Outer evaluation uses stratified 5-fold cross-validation (classification) or standard 5-fold (regression).

\paragraph{Activation baseline.}
The last-token hidden state of the complete response at the monitored layer is extracted for each sample, yielding a vector of dimension $d$ (the model's hidden size). PCA is applied to reduce this to 128 components before fitting the same regularized classifier, preserving as much variance as possible while keeping the problem tractable.

\paragraph{Random-projection baseline.}
Eight random unit vectors are drawn once and fixed; activation projections onto these vectors are computed using the same feature extraction pipeline as polylogue. By the Johnson–Lindenstrauss lemma, random projections approximately preserve pairwise distances, so this baseline tests whether \emph{any} low-dimensional summary of activation geometry is predictive, not specifically the interpretable persona directions.

\subsection{All Coefficients}
\label{app:all_coefficients}

Table~\ref{tab:correctness_coefficients} shows the top-5 features per model. The top features differ across models, suggesting that latent dynamics linked to correctness are not shared across models. For \emph{Qwen2.5-14B-Instruct}, volatility features (\emph{solver}, \emph{analyst}, \emph{explorer}) and early \emph{verifier} suppression dominate, so variability across the trace matters more than the final state. For \emph{DeepSeek-R1-Distill-Qwen-14B}, the strongest predictor is final-step \emph{monitor} ($-0.74$): still monitoring at the end is associated with wrong answers. For \emph{Phi-4-reasoning}, four of the top five features sit near the answer, with final-step \emph{analyst} hurting and \emph{arbiter} helping. For \emph{Llama-3.1-Nemotron-Nano-8B-v1}, \emph{explorer} volatility and final-step \emph{monitor}/\emph{analyst} all carry large negative weights, linking erratic exploration and unresolved late analysis to failure.

A clear split appears between reasoning and non-reasoning models. For the three reasoning models, final-step features (\emph{monitor}, \emph{analyst}, \emph{arbiter}) carry the largest weights, so the state at answer commitment is most predictive of correctness. For the non-reasoning Qwen2.5-14B-Instruct, no final-step feature appears in the top-5; predictive signal is spread over volatility and mid-trace paragraphs instead.

\begin{table*}[ht]
\scriptsize
\centering
\begin{tabular}{rlrlrlrlr}
\toprule
Rank & \multicolumn{2}{c}{Qwen2.5-14B-Instruct} & \multicolumn{2}{c}{DeepSeek-R1-Distill-Qwen-14B} & \multicolumn{2}{c}{Phi-4-reasoning} & \multicolumn{2}{c}{Llama-3.1-Nemotron-Nano-8B-v1} \\
\cmidrule(lr){2-3} \cmidrule(lr){4-5} \cmidrule(lr){6-7} \cmidrule(lr){8-9}
& Feature & Coef. & Feature & Coef. & Feature & Coef. & Feature & Coef. \\
\midrule
1 & para 1 verifier & -0.39 & final sim monitor & -0.74 & final sim analyst & -0.68 & volatility explorer & -0.53 \\
2 & para 4 explorer & +0.36 & para 0 interpreter & +0.16 & final sim arbiter & +0.36 & final sim monitor & -0.51 \\
3 & volatility solver & +0.36 & para 7 interpreter & -0.12 & dominance share arbiter & +0.34 & final sim analyst & -0.26 \\
4 & volatility analyst & +0.36 & para 19 planner & +0.12 & para 19 arbiter & +0.28 & volatility planner & +0.25 \\
5 & volatility explorer & -0.31 & para 18 arbiter & +0.12 & volatility solver & +0.17 & para 3 explorer & -0.19 \\
\bottomrule
\end{tabular}
\caption{Top-5 coefficients of the L1 logistic regression on polylogue features per model. \textit{para $p$ persona}: mean alignment of \textit{persona} over paragraph $p$. \textit{final sim persona}: alignment at the final generation step. \textit{volatility persona}: standard deviation of alignment across all steps. \textit{dominance share persona}: fraction of steps at which \textit{persona} is the dominant (highest-scoring) direction.}
\label{tab:correctness_coefficients}
\end{table*}

\section{Causal Faithfulness}
\label{app:causal_faithfulness}

\subsection{Steering strategy derivation}
After fitting the functional faithfulness model, we extract the top-5 non-zero Lasso coefficients ranked by absolute value. Each coefficient corresponds to a (paragraph index, persona) pair. We map paragraph indices to 1-indexed paragraph ranges using the median paragraph count of the training responses. Positive coefficients yield a steering direction of $+1$ (amplify the trait); negative coefficients yield $-1$ (suppress the trait).

\subsection{Steering mechanism}
Steering is applied via three components:
\begin{itemize}
  \item \textbf{MultiDynamicActivationSteerer}: registers a forward hook on the specified transformer layer and, at each generation step, adds $\alpha \cdot v_k$ to the last-token hidden state for each active trait $k$, where $\alpha$ is the per-model steering coefficient and $v_k$ the persona vector.
  \item \textbf{ParagraphJudge}: tracks the number of double-newline separators decoded so far to determine the current paragraph number for each sequence in the batch, and returns a Boolean activation mask per trait based on the steering strategy.
  \item \textbf{MultiSteeringProcessor}: a \textbf{LogitsProcessor} called at each generation step that queries the \textbf{ParagraphJudge} and writes the resulting masks into the steerer before the forward pass.
\end{itemize}

\section{Computational Resources}
\label{app:compute}

Table~\ref{tab:model_sizes} lists the parameter count of each evaluated model and the LLM judge (Llama-3.3-70B-Instruct, Appendix~\ref{app:selecting}, Section~\ref{sec:semantic_faithfulness}).

\begin{table}[h]
\footnotesize
\centering
\begin{tabular}{lr}
\toprule
Model & Params \\
\midrule
Qwen2.5-14B-Instruct & 14.7B \\
DeepSeek-R1-Distill-Qwen-14B & 14.7B \\
Phi-4-reasoning & 14.7B \\
Llama-3.1-Nemotron-Nano-8B-v1 & 8.03B \\
Llama-3.3-70B-Instruct (judge) & 70B \\
\bottomrule
\end{tabular}
\caption{Parameter count for each evaluated model and the LLM judge used for scoring and tagging.}
\label{tab:model_sizes}
\end{table}

All experiments ran on a SLURM-managed cluster on NVIDIA A100 (80GB) and H100 (94GB) GPUs (PyTorch 2.6.0, vLLM 0.8.5.post1, CUDA 12.4), using 1--2 GPUs per job. Total compute across all experiments was approximately \textbf{105 GPU-hours}.\footnote{Estimated from per-job timing logged in the pipeline scripts (persona vector extraction and judging, layer/coefficient selection, MMLU-Pro generation, and steering); excludes CPU-only fitting/analysis jobs.}

\end{document}